%% file: manuscript-amspreprint.tex
\pdfoutput=1

\IfFileExists{./prepreamble-amspreprint.sty}{\RequirePackage[packages,theorems,changes]{./prepreamble-amspreprint}}{}

\makeatletter
\IfFileExists{./scoop-latex/scoop-packages.sty}{\providecommand*{\input@path}{}\edef\input@path{{./scoop-latex/}\input@path}}{}
\makeatother
\PassOptionsToPackage{sorting = none, style = nature, backend = biber}{biblatex}
\PassOptionsToPackage{commentmarkup = footnote}{changes}    

\documentclass[english]{amsart}

\RequirePackage{biblatex}
\RequirePackage{ltxcmds}
\IfFileExists{./preamble-amspreprint.sty}{\RequirePackage[packages,theorems,changes]{./preamble-amspreprint}}{}

\usepackage{manuscript}

\IfFileExists{./postpreamble-amspreprint.sty}{\RequirePackage[packages,theorems,changes]{./postpreamble-amspreprint}}{}

\makeatletter
\@ifpackageloaded{changes}{
\definechangesauthor[name = {Viktor Martinek}, color = {red!80!black}]{VM}
\definechangesauthor[name = {Julia Reuter}, color = {blue!80!black}]{JR}
\definechangesauthor[name = {Ophelia Frotscher}, color = {orange!80!black}]{OF}
\definechangesauthor[name = {Sanaz Mostaghim}, color = {green!70!black}]{SM}
\definechangesauthor[name = {Markus Richter}, color = {teal}]{MR}
\definechangesauthor[name = {Roland Herzog}, color = {red!80!black}]{RH}
}{}
\makeatother        

\makeatletter
\@ifpackageloaded{hyperref}{%
	\hypersetup{
		pdftitle = {Shape Constraints in Symbolic Regression using Penalized Least Squares},
		pdfauthor = {Viktor Martinek, Julia Reuter, Ophelia Frotscher, Sanaz Mostaghim, Markus Richter, Roland Herzog},
		pdfkeywords = {symbolic regression, shape constraints, constrained least squares, thermodynamics, equations of state},
		pdfcreator = {Created using the Scoop Template Engine version 1.2.0.post14+git.c1c7e129.dirty.}
	}
}{
	\pdfinfo{
		/Title (Shape Constraints in Symbolic Regression using Penalized Least Squares)
		/Author (Viktor Martinek, Julia Reuter, Ophelia Frotscher, Sanaz Mostaghim, Markus Richter, Roland Herzog)
		/Subject ()
		/Keywords (symbolic regression, shape constraints, constrained least squares, thermodynamics, equations of state)
		/Creator (Created using the Scoop Template Engine version 1.2.0.post14+git.c1c7e129.dirty.)
	}
}
\makeatother

\title[Shape Constraints in Symbolic Regression]{Shape Constraints in Symbolic Regression using Penalized Least Squares}

\author[V. Martinek]{Viktor Martinek\orcidlink{0000-0001-6215-4783}}
\address[V. Martinek]{Interdisciplinary Center for Scientific Computing, Heidelberg University, 69120 Heidelberg, Germany}
\email{viktor.martinek@iwr.uni-heidelberg.de}

\author[J. Reuter]{Julia Reuter\orcidlink{0000-0002-7023-7965}}
\address[J. Reuter]{Otto von Guericke University, Faculty of Computer Science, Chair of Computational Intelligence, 39106 Magdeburg, Germany}
\email{julia.reuter@ovgu.de}

\author[O. Frotscher]{Ophelia Frotscher\orcidlink{0000-0002-6915-1988}}
\address[O. Frotscher]{Faculty of Mechanical Engineering, University of Technology Chemnitz, Applied Thermodynamics, 09107 Chemnitz, Germany}
\email{ophelia.frotscher@mb.tu-chemnitz.de}

\author[S. Mostaghim]{Sanaz Mostaghim\orcidlink{0000-0002-9917-5227}}
\address[S. Mostaghim]{Otto von Guericke University, Faculty of Computer Science, Chair of Computational Intelligence, 39106 Magdeburg, Germany}
\email{sanaz.mostaghim@ovgu.de}

\author[M. Richter]{Markus Richter\orcidlink{0000-0001-8120-5646}}
\address[M. Richter]{Faculty of Mechanical Engineering, University of Technology Chemnitz, Applied Thermodynamics, 09107 Chemnitz, Germany}
\email{m.richter@mb.tu-chemnitz.de}

\author[R. Herzog]{Roland Herzog\orcidlink{0000-0003-2164-6575}}
\address[R. Herzog]{Interdisciplinary Center for Scientific Computing, Heidelberg University, 69120 Heidelberg, Germany}
\email{roland.herzog@iwr.uni-heidelberg.de}

\thanks{This work was funded by the Deutsche Forschungsgemeinschaft (DFG, German Research Foundation) -- HE~6077/14-1 and RI~2482/10-1 -- within the Priority Programme \enquote{SPP 2331: Machine Learning in Chemical Engineering}.}

\date{\today}

\dedicatory{}

\begin{document}

\begin{abstract}
\input{abstract.tex}
\end{abstract}


\makeatletter
\ltx@ifpackageloaded{hyperref}{%
\subjclass[2010]{}
}{%
\subjclass[2010]{}
}
\makeatother

\maketitle

\input{content.tex}

\appendix

\printbibliography

\end{document}

%% file: abstract.tex
We study the addition of shape constraints (SC) and their consideration during the parameter identification step of symbolic regression (SR).
SC serve as a means to introduce prior knowledge about the shape of the otherwise unknown model function into SR.
Unlike previous works that have explored SC in SR, we propose minimizing SC violations during parameter identification using gradient-based numerical optimization.
We test three algorithm variants to evaluate their performance in identifying three symbolic expressions from synthetically generated data sets.
This paper examines two benchmark scenarios: one with varying noise levels and another with reduced amounts of training data.
The results indicate that incorporating SC into the expression search is particularly beneficial when data is scarce.
Compared to using SC only in the selection process, our approach of minimizing violations during parameter identification shows a statistically significant benefit in some of our test cases, without being significantly worse in any instance.

%% file: content.tex
\section{Introduction}
\label{section:introduction}

\Ac{SR} is a supervised machine learning approach to discover underlying mathematical expressions from data.
The search space of potential expressions is very large, and the problem is NP-hard, as proven by \cite{VirgolinPissis:2022:1}.
However, there are ways to incorporate prior knowledge beyond the data into the expression search to guide the search more efficiently.
One way to do this is by using \ac{SC}.

\ac{SC} impose restrictions on the behavior (or shape) of a function.
A relatively common example of a \ac{SC} is monotonicity with respect to some variable.
\ac{SC} can be used to incorporate domain-specific knowledge or enforce desired behavior, which can be advantageous in many scenarios.
For example, \ac{SC} can help reduce the required number of data, or help bridge regions where no data is available.
In addition, \ac{SC} can  mitigate the effect of noisy data, including data with outliers.
Depending on how \ac{SC} are implemented into the expression search, they can either provide additional guidance or even reduce the search space.
These benefits are already shown in several studies, including \cite{KronbergerOlivettiDeFrancaBurlacuHaiderKommenda:2022:1,HaiderOlivettiDeFrancaBurlacuKronberger:2021:1,HaiderOlivettiDeFrancaKronbergerBurlacu:2022:1,HaiderKronberger:2022:1,BladekKrawiec:2023:1,KubalikDernerBabuska:2020:1,PiringerWagnerHaiderFohlerSilberAffenzeller:2022:1,ReinboldKageorgeSchatzGrigoriev:2021:1,LiFanSinghRiley:2019:1}.
Altogether, the expressions obtained have greater utility and reliability when they obey the domain-specific \ac{SC}.

Our proposed approach differs from previous work in two points.
First, we take into account the amount of \ac{SC} violations already during the parameter identification step.
To the best of our knowledge, this is only done in one other publication (see \cite{KubalikDernerBabuska:2020:1}).
Second, this parameter identification step is carried out using second-order optimization and algorithmic differentiation.
This allows us to find a set of parameters such that the resulting model conforms to the \ac{SC}, while also fitting the data.
As in other works, the \ac{SC} violations are implemented as additional objectives which are considered in the multi-objective selection process of the genetic algorithm.

In the following section, we introduce the problem formulation of \ac{SR} and extend it to include \ac{SC}.
We describe our approach to the parameter identification, which also minimizes the \ac{SC} violations.
We then explain how we incorporate the \ac{SC} into the evolutionary algorithm.
In \cref{section:relatedwork}, related works are briefly described and the differences to the present work are outlined.
In \cref{section:experiments}, the experiments are detailed, and their results presented.
\Cref{section:discussion} offers a summary and discussion of the results.
Finally, we draw a conclusion in \cref{section:conclusion}.

\section{Shape Constraints in Symbolic Regression}
\label{section:shape-constraints-in-SR}

In \ac{SR}, the goal is to find a mathematical expression~$m$, along with values for its parameters~$p$, such that we
\begin{equation}
	\label{eq:SR-problem}
	\text{Minimize}
	\quad
	\frac{1}{N} \sum_{i=1}^N f \paren[bigg](){\frac{y_i - m(X_i, p)}{y_i}}
	\quad
	\text{\wrt\ }
	(m,p)
	.
\end{equation}
The expression~$m$ comes from an admissible class of expressions ensuring its syntactic correctness.
The parameter~$p$ is from a parameter space that generally depends on~$m$.
Above, the vector-valued $X$ and the scalar-valued $y$ are the independent and dependent variables of the model, and $(X_i,y_i)$ denote the pairs of data in the data set, $i = 1, \ldots, N$.
The function~$f$ is known as loss function and typical examples include $f(r) = r^2$, where $r$ may be the residual resulting from $y_i - m(X_i, p)$.
This example leads to \eqref{eq:SR-problem} representing the mean squared relative error.

Currently, genetic programming algorithms perform best among wide-ranging \ac{SR} tasks (see \cite{LaCavaOrzechowskiBurlacuOlivettiDeFrancaVirgolinJinKommendaMoore:2021:1}).
Summarized briefly, the population-based genetic programming algorithms combine candidate expressions (crossover), and insert slight variations into candidate expressions (mutation).
In this way, new candidate expressions are generated.
Better candidate models replace less favorable ones in the population, which is known as the selection process.

As shorter and intuitively simpler expressions are preferred, problem \eqref{eq:SR-problem} is augmented by an additional objective measuring the complexity, \eg, length, of the expression~$m$ in most \ac{SR} approaches.
Many approaches consider additional objectives, such as age (see \cite{SchmidtLipson:2010:1}), or the coefficient of determination $R^2$.

In any case, \eqref{eq:SR-problem} becomes a multi-objective problem.
Depending on the algorithm, these competing objectives are either addressed through the concepts of Pareto dominance, or aggregated to a yield singular fitness value by means of a weighted sum.

In this work, we use \ac{TiSR} (see \cite{MartinekFrotscherRichterHerzog:2023:1}), an \ac{SR} implementation based on NSGA-II (see \cite{DebPratapAgarwalMeyarivan:2002:1}).
In contrast to NSGA-II, \ac{TiSR} uses both Pareto dominance and fitness-based selection criteria to ensure sufficient diversity in the population.
\ac{TiSR} is available at \url{https://github.com/scoop-group/TiSR}.

\ac{SC} are expressed by augmenting problem \eqref{eq:SR-problem} using inequality and/or equality constraints.
The problem then becomes
\begin{equation}
	\begin{aligned}
		\text{Minimize}
		\quad
		&
		\frac{1}{N} \sum_{i=1}^N f \paren[bigg](){\frac{y_i - m(X_i, p)}{y_i}}
		\quad
		\text{\wrt\ }
		(m,p)
		\\
		\text{subject to}
		\quad
		&
		g(m(X, p)) \le 0
		\quad
		\text{for all }
		X \in \Omega
		,
		\\
		\text{and}
		\quad
		&
		h(m(X, p)) = 0
		\quad
		\text{for all }
		X \in \Omega
		.
	\end{aligned}
	\label{eq:SR-problem-with-shape-constraints}
\end{equation}
The set $\Omega$ defines the range for the independent variable~$X$ on which the \ac{SC} is relevant.
Therefore, in general, there are infinitely many \ac{SC}, that can be taken into account in two different ways.

In pessimistic approaches, interval arithmetic is used to compute an enclosure of the range of the \ac{SC} function's values over $X \in \Omega$.
This offers algorithms the possibility to certify a model's feasibility \wrt all \ac{SC}.
However, such approaches may deem feasible models infeasible, as observed and discussed by Haider~et~al.\ (see \cite{HaiderOlivettiDeFrancaKronbergerBurlacu:2022:1}).
By contrast, optimistic approaches evaluate and enforce the \ac{SC} only in a finite number of predetermined locations.
While this may lead to models incorrectly classified as feasible, the evaluation of \ac{SC} as well as their derivatives \wrt $p$ is straightforward.

In this study, we employ an optimistic approach.
The \ac{SC} evaluation points $X^1, \ldots, X^C$ are distinct from the data points $X_1, \ldots, X_N$.
In order to accommodate different types of \ac{SC}, the \ac{SC} functions~$g$ and~$h$ can be vector-valued.
This allows several \ac{SC} to be considered simultaneously, \eg, monotonicity and bounds on the function values.
For example, to realize a monotonicity \ac{SC}, $g(m(X, p))$ could calculate the first derivatives of $m$ on all points $X$ with the paramters $p$.

The parameters~$p$ of candidate model~$m$ can be identified using the evolutionary mechanisms only, as exemplified by Koza in the early literature on genetic programming for \ac{SR} (see \cite{Koza:1994:1}). 
Using this approach, no separate parameter identification step with a fixed model is conducted, and only mutation and selection are used to identify the model along with its parameters~$m(p)$.
However, most contemporary \ac{SR} algorithms treat the parameter identification as a subordinate problem, employing gradient-based fitting for each candidate model before the loss in \eqref{eq:SR-problem} is determined.
Kommenda~et~al.\ have shown the benefits associated with this approach (see \cite{Kommenda:2018:1}).

Using $f(r) = r^2$, the parameter identification subproblem, \ie, \eqref{eq:SR-problem-with-shape-constraints} with the model~$m$ fixed, becomes a constrained least-squares problem.
Despite the relevance of this problem class in many industrial and academic fields, we were unable to find an appropriate open-source implementation of a constrained least-squares problem solver.

We therefore employ a penalty approach.
Given a model~$m$ with parameters~$p$, we measure the total violation of the \ac{SC} in terms of the $\ell_2$-squared penalty, \ie,
\begin{equation}
	\label{eq:constraint-violation}
	\sum_k \sum_{\ell=1}^C h_k(m(X^\ell, p))^2
	\quad
	\text{and}
	\quad
	\sum_k \sum_{\ell=1}^C \max \paren[big](){0, g_k(m(X^\ell, p))}^2
	,
\end{equation}
where $k$ are the different \ac{SC} and $C$ are the points they are evaluated on.
These values represent the extent to which the \ac{SC} are violated at the points of evaluation.
Notice that the \ac{SC} evaluation points $X^1, \ldots, X^C$ are the same for each \ac{SC} in \eqref{eq:constraint-violation}.
This is merely for the sake of notational convenience.
In our implementation, we allow the \ac{SC} evaluation points to be different for each component $h_k$ and $g_k$ of the \ac{SC} functions.

During the parameter identification step in our algorithm, we minimize a weighted sum of the loss and the \ac{SC} violation \eqref{eq:constraint-violation}.
In order to keep the computational expense limited, we work with a fixed penalty parameter.
We use the \namemd{Newton} implementation of the \julia library \namemd{Optim.jl} (see \cite{MogensenRiseth:2018:1}) for this purpose.
First- and second-order derivatives of the loss as well as the \ac{SC} are provided through the algorithmic differentiation package \namemd{ForwardDiff.jl} (see \cite{RevelsLubinPapamarkou:2016:1}).
This is an essential distinction from \cite{KubalikDernerBabuska:2020:1}, where Monte-Carlo-based multi-objective local search is used for parameter fitting.
Gradient-based approaches, where available, often outperform non-deterministic approaches (see \cite{MartinsNing:2021:1}).

With the parameter values optimized, there are at least two distinct ways for an evolutionary algorithm to take into account the amount of \ac{SC} violation encountered in a particular model.
In so-called hard-constrained approaches, candidate models are removed from the population in case they are infeasible.
By contrast, soft-constrained approaches incorporate \ac{SC} violations through one or several additional objectives.
Hard constraints may lead to a loss of diversity in the population, which is essential for evolutionary algorithms, while soft \ac{SC} allow to improve upon the candidate in later generations.

Haider~et~al.\ compared hard- and soft-constrained approaches and found no statistically significant differences between them (see \cite{HaiderOlivettiDeFrancaBurlacuKronberger:2021:1}).
We therefore choose to implement a soft-constrained approach, where the two terms in \eqref{eq:constraint-violation} are added to form a single additional objective.

\section{Related Work}
\label{section:relatedwork}

In an early work, \cite{LiFanSinghRiley:2019:1} introduce asymptotic \ac{SC} into \ac{SR}, controlling the function behavior at zero and infinity.
They utilize a neural network to guide their Monte-Carlo tree search-based method and to produce expressions with the desired asymptotic behavior.
Although effective through restriction of the search space, it is not directly applicable to other types of \ac{SC}.

\cite{BladekKrawiec:2019:1,BladekKrawiec:2023:1} use a satisfiability modulo theories (SMT).
The SMT solver guarantees that solutions are feasible, resembling a pessimistic approach.
The types of \ac{SC} considered include bounds on function values and partial derivatives, as well as symmetries and periodicity.
Their approach, namely \enquote{Counterexample-Driven Genetic Programming}, also maintains a training set with counter examples, which is continuously extended during the expression search.
However, SMT solvers may be computationally prohibitive and do not support all types of \ac{SC}.

Closest to our approach in the present work is \cite{KubalikDernerBabuska:2020:1}.
The \ac{SC} are evaluated on a finite number of sample points.
Similar to our approach, the \ac{SC} are part of the parameter identification subproblem.
Contrary, however, they use a multi-objective local search procedure based on Monte-Carlo methods for this problem.
This avoids computing derivatives with respect to the parameters, but gradient-based approaches often outperform non-deterministic approaches (see \cite{MartinsNing:2021:1}).

\cite{KronbergerOlivettiDeFrancaBurlacuHaiderKommenda:2022:1} introduce a new approach to shape-constrained \ac{SR}.
They use interval arithmetic to check \ac{SC} satisfaction in a pessimistic manner.
Their study is limited to box constraints on the function values and its partial derivatives.
Two approaches of incorporating the \ac{SC} into the expression search are studied.
One uses hard constraints, while the other maintains separate populations for feasible and infeasible candidates.
In constrast to our work, the \ac{SC} violations are not part of the parameter identification subproblem.

\cite{HaiderOlivettiDeFrancaBurlacuKronberger:2021:1} employs soft constraints which allows the \ac{SC} violation to be reduced over multiple generations.
The authors utilize three different evolutionary algorithms as a basis and study various ways of treating and incorporating \ac{SC} into the expression search.
The benefits of \ac{SC}, \eg, for noisy data, are clearly shown.
Interestingly, the authors conclude that in all their experiments, there are no statistically significant differences between the core algorithms nor among the approaches to incorporate the \ac{SC}.

\cite{HaiderOlivettiDeFrancaKronbergerBurlacu:2022:1} compare optimistic and pessimistic \ac{SC} evaluation.
The authors conclude that optimistic approaches improve the extrapolation behavior of models, while pessimistic approaches are better suited for higher levels of noise.

\section{Experiments}
\label{section:experiments}

We conduct experiments on three different benchmark problems.
In each case, a ground truth model is known and used to generate synthetic data.
The goal of our experiments is to study the utility in using \ac{SC} to reconstruct the ground truth model.
In contrast to previous works, we do not emphasize the effects of \ac{SC} on the composition of the final Pareto frontier (hall of fame).

In what follows, we first detail how we set up our proposed method and two baseline methods we compare our method to.
This is followed by a description of the three benchmark problems.

\subsection{Description of Algorithmic Variants}

\begin{figure}[hbt]
    \centering
	\includegraphics[width=0.9\textwidth]{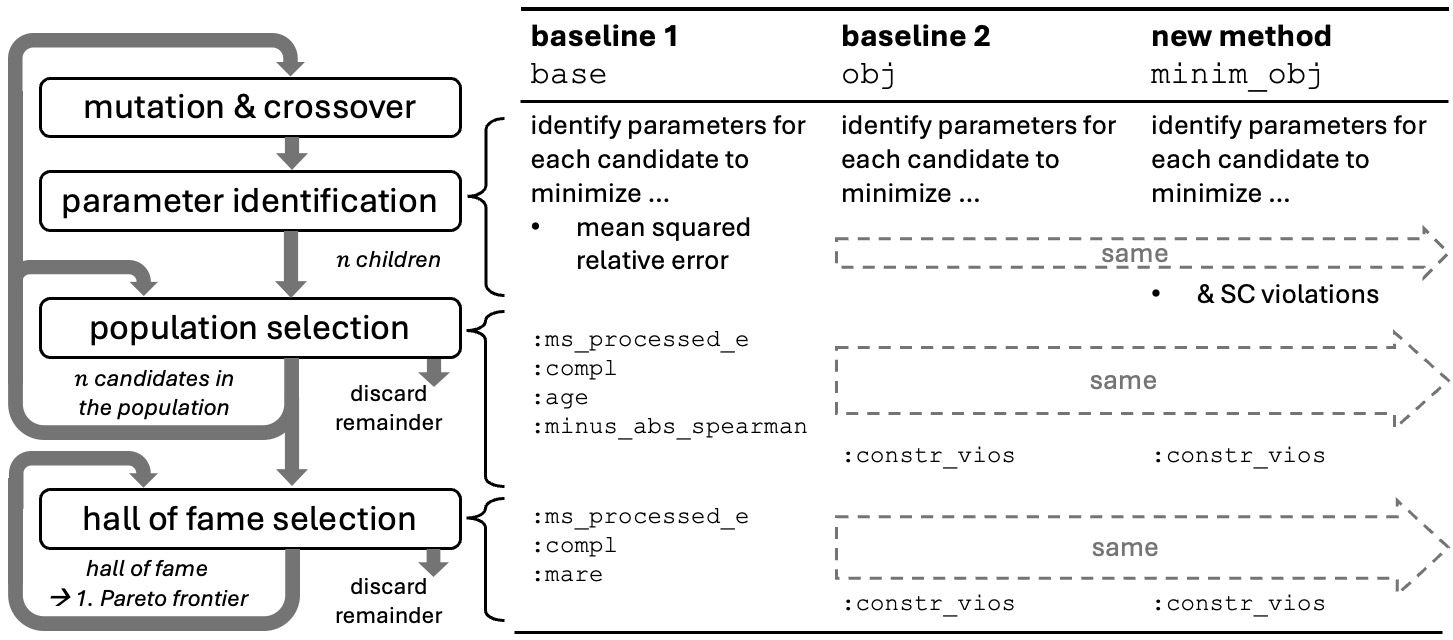}
	\caption{Simplified illustration of NSGA-II (left) and overview of the key differences between the three algorithm variants with regard to parameter identification, population selection, and hall of fame selection (right).}
	\label{fig:overview_figure}
\end{figure}

All variants are implemented in \ac{TiSR}.
\begin{enumeratearabic*}
	\item
		As a first baseline (\base), we use \ac{TiSR} without \ac{SC}.

	\item
		The second baseline (\obj) takes \ac{SC} into account partially.
        As in \base, it solves the parameter identification problem without considering the \ac{SC}.
		It then evaluates the \ac{SC} violation \eqref{eq:constraint-violation} and introduces the combined amount of violation as a separate, additional objective.
		This approach resembles the one of Haider~et~al.\ (see \cite{HaiderOlivettiDeFrancaBurlacuKronberger:2021:1}).

	\item
		The third variant (\minimobj) is the one proposed in this work.
        It alters the parameter identification step.
		To lower the computational expense, it first identifies the parameters them without considering the \ac{SC}, like both baseline methods.
		If certain criteria are met (mentioned below), the parameter identification is continued using the penalty-based constrained least squares.
		The \ac{SC} violations after the parameter identification are used for the selection, like in \obj.
\end{enumeratearabic*}
The key differences are summarized in \cref{fig:overview_figure}.

Most of \ac{TiSR}'s default parameters are used in all three variants.
To minimize the mean squared relative error, as shown in \eqref{eq:SR-problem}, the \parameter{fit_weights} parameter is set to $1 / y_i$ for each residual value $r_i$, where $y_i$ is the dependent variable for datum~$i$.
The maximum complexity, \ie, the number of operators and operands, is set to 5~above the reference ground truth expression.
The default selection objectives of \ac{TiSR} \parameter{[:ms_processed_e, :compl, :age]} are the mean squared error, complexity of the expression, and age of the expression (number of generations it has been present in the population).
For the purpose of the present study, we extend these objectives by \parameter{:minus_abs_spear} and \parameter{:constr_vios}.
Here \parameter{:minus_abs_spear} is the negative of the absolute of the Spearman correlation, which is added as a supporting objective as defined in \cite{ZilleEvrardReuterMostaghimVanWachem:2021:1}.
The \parameter{:constr_vios} objective contains the sum of the \ac{SC} violations as in \eqref{eq:constraint-violation}, which is always zero for the \base variant of the algorithm.
The default hall of fame objectives \parameter{[:ms_processed_e, :compl, :mare]} are extended by \parameter{:constr_vios}.
The \parameter{:mare} objective represents the \ac{MARE}, which is calculated as in \eqref{eq:SR-problem}, but with loss function $f(r) = \abs{r}$ in place of $f(r) = r^2$.
Further parameters of \ac{TiSR} that deviate from their defaults are summarized in \Cref{table:params}.

\begin{table}[htb]
	\centering
	\caption{Overview and description of some of \ac{TiSR}'s parameters set for all experiments deviating from their default values.}
	\begin{tabular}{@{}p{0.32\textwidth}@{}p{0.11\textwidth}@{}p{0.57\textwidth}}
		\toprule
		\textbf{Parameter}                  & \textbf{Value}      & \textbf{Description}
		\\
		\midrule
		\parameter{t_lim}                   & \parameter{60 * 10} & The total runtime is set to ten minutes.
		\\
		\parameter{pop_size}                & \parameter{500}     & The population size, \ie, the number of candidate models in the population, is set to 500.
		\\
		\parameter{pow_abs_param}           & \parameter{true}    & The power operator (\parameter{^}) can only have parameters in its exponent. Variables or expressions are not allowed as exponents.
		\\
		\parameter{always_drastic_simplify} & \parameter{1e-7}    & Candidate models containing parameters that are closer to one or zero than $10^{-7}$ are simplified by rounding the respective parameters to one or zero.
		\\
		\bottomrule
	\end{tabular}
	\label{table:params}
\end{table}

All three algorithmic variants are allowed to use up to 30~iterations for each parameter identification subproblem.
In fact, the \base and \obj variants use a Levenberg-Marquardt method for this purpose.
The third variant \minimobj allots a budget of up to 20~iterations without \ac{SC} penalization.
Provided that the \ac{MARE} is smaller than a pre-selected boundary, an additional maximum of 10~iterations with \ac{SC} penalization follow.
This boundary is set \SI[round-mode = figures, round-precision = 1]{5}{\percent} higher than the set noise level of the data.
The \ac{SC} penalty coefficient is fixed at~$1$.

\subsection{Verification of Results}

To check whether the ground truth is reconstructed, a separate verification data set is used.
This data set is sampled on a wider range than the original data set, and using 2~to~5~times as much data.
Approximately every $\SI[round-mode = figures, round-precision = 1]{5}{\second}$, all candidate solutions in the hall of fame are passed to a callback function.
Those with a \ac{MARE} of at most \SI[round-mode = figures, round-precision = 1]{5}{\percent} higher than the noise level of the data are fitted to the verification data set.
If the \ac{MARE} with respect to the verification data set is lower than \SI[round-mode = figures, round-precision = 1]{0.0001}{\percent}, the experiment is terminated with success.
We recall that only expressions with a complexity of up to 5~above the reference ground truth expression can be accepted.
The expression search does not gain any information from the verification data set or the callback overall.

\subsection{Description of the Benchmark Problems}

We report on three different problems, each using a different ground truth expression.
In all cases, no part of the expression nor its parameters are initially known to \ac{TiSR}.
Every problem has two independent and one dependent variable.

\subsubsection{Gaussian Distribution}

\begin{figure}[htb]
	\begin{subfigure}[t]{0.48\textwidth}
		\includegraphics[width = \linewidth]{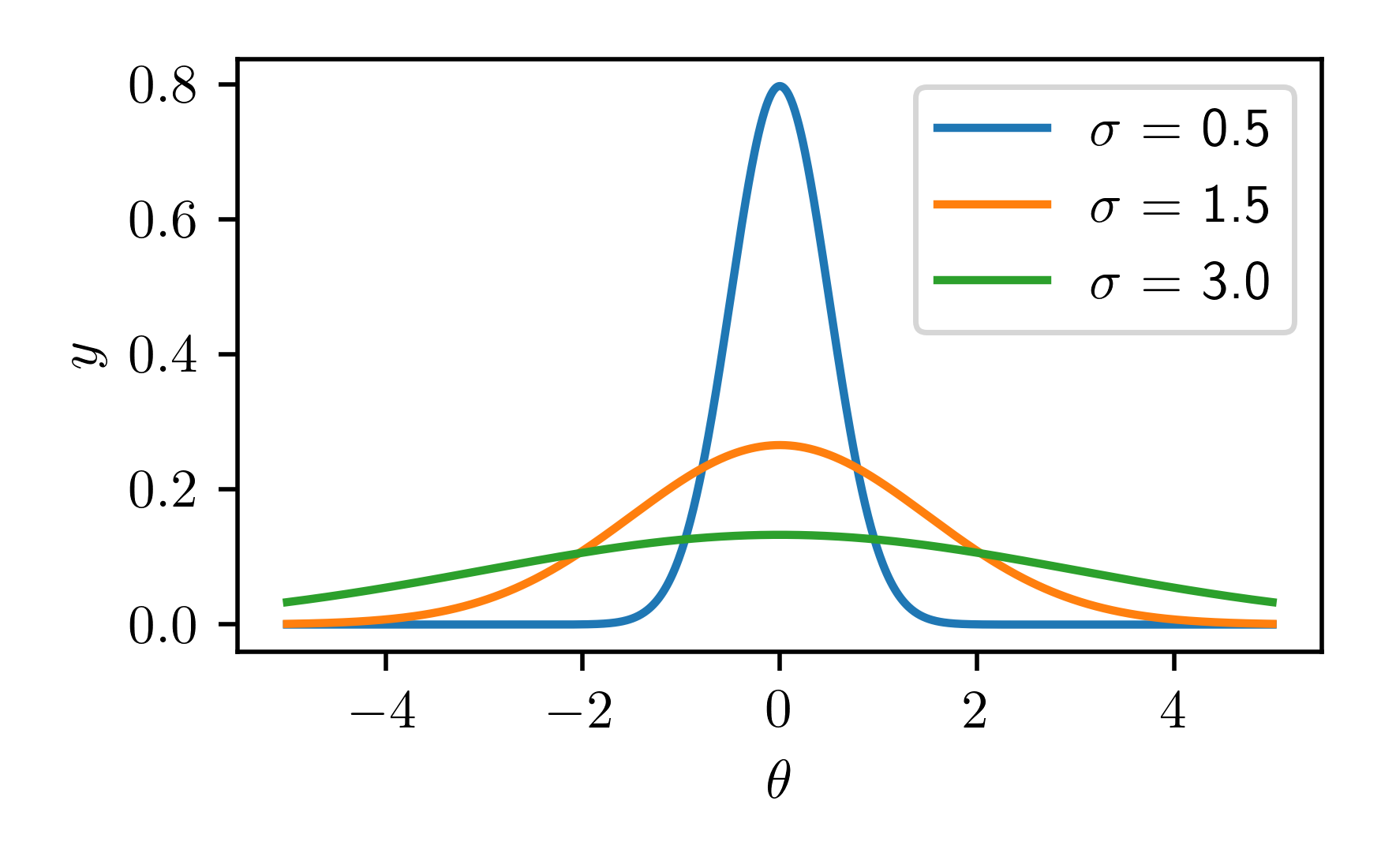}
	\end{subfigure}
	\hfill
	\begin{subfigure}[t]{0.48\textwidth}
		\includegraphics[width = \linewidth]{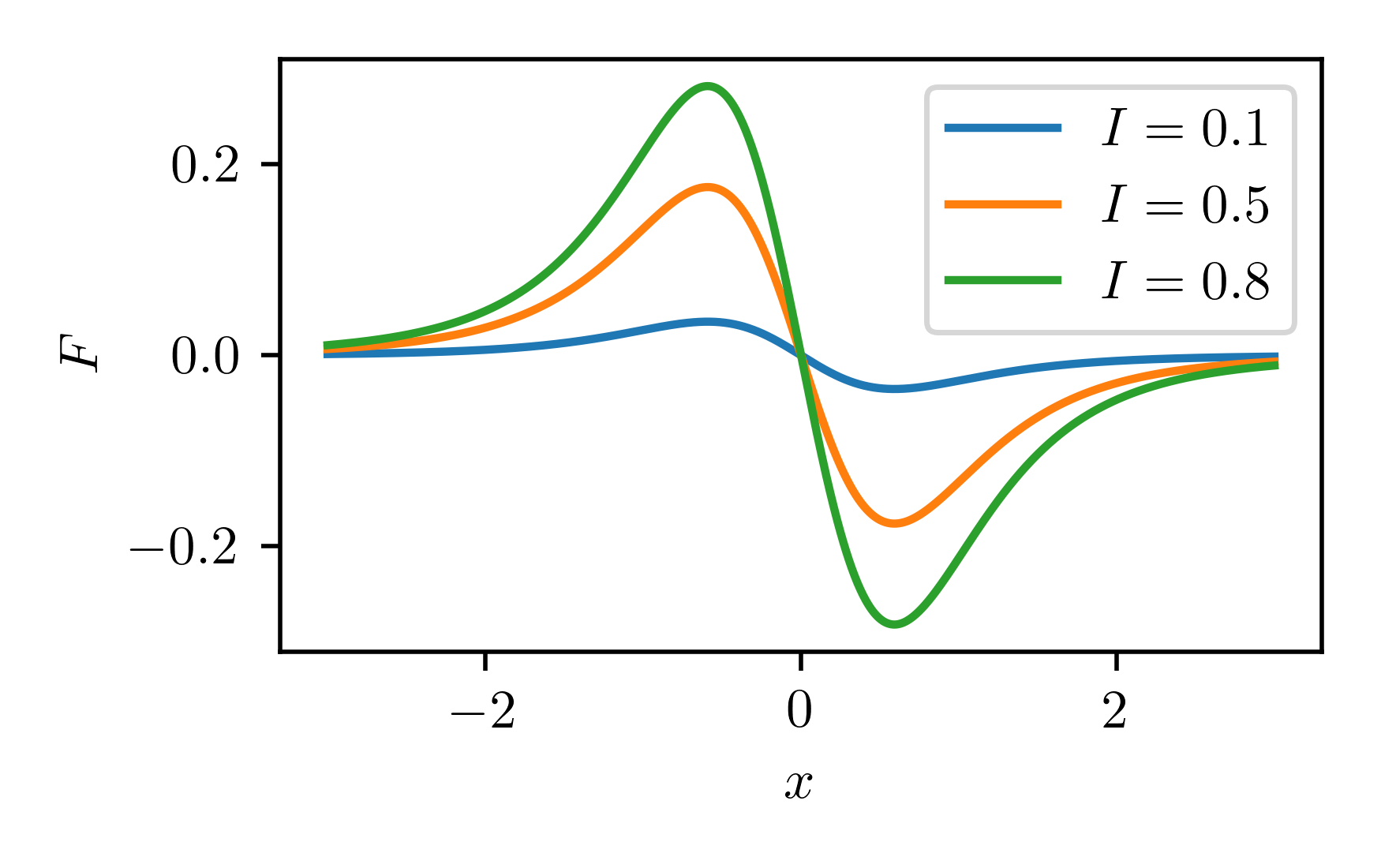}
	\end{subfigure}
	\caption{Plot of the \gaussian expression for varying standard deviation $\sigma$ and random variable values $\theta$ (left) and the \magman expression for varying distances~$x$ and varying current~$I$ (right).}
	\label{fig:gaussianfunc}
	\label{fig:magmanfunc}
\end{figure}

The first problem (\gaussian) is the density of a univariate Gaussian distribution with the mean fixed at~0:
\begin{equation}
	y
	=
	\frac{\exp \paren[big](){-\frac{\theta^2}{2 \sigma^2}}}{\sqrt{2 \pi} \, \sigma}
	.
	\label{equation:gaussian}
\end{equation}
The independent variables~$\sigma$ and $\theta$ denote the standard deviation and the probability density, respectively.
A graph illustrating \cref{equation:gaussian} is shown in \Cref{fig:gaussianfunc} on the left.
We choose this as a benchmark problem, as there are many \ac{SC} possible.
The ground truth expression has a complexity of 11.

For all variables, ranges are defined and within those, 100 points uniformly, randomly sampled.
The ranges for the fitting data set are $[-5, 5]$ for $\theta$, and $[0.5, 3]$ for $\sigma$.
For the verification data set, 500~points are sampled within the ranges of $[-10, 10]$ and $[0.5, 5]$ for $\theta$ and $\sigma$, respectively.
The function set for this expression is set to \parameter{[+,  -,  *,  /,  ^, exp, pow2, sqrt]}, where \parameter{pow2} is a unary function raising to the power of~2.

We define \ac{SC} on the function values and the first derivatives of the function.
The \ac{SC} on the function value are summarized in \Cref{table:gaussiandomaincons}, while the \ac{SC} on the first derivatives are listed in \Cref{table:gaussianmonocons}.
For each of those, five points are sampled for each variable individually.
Then five~$(\theta,\sigma)$ pairs are randomly drawn from the total of the 25~potential pairs.
These \ac{SC} evaluation points remain the same during each run but may differ between runs.

\begin{table}[htb]
	\caption{Collection of tables detailing the treatment of \ac{SC} for \gaussian and the \magman expression.}
	\centering
	\begin{subtable}{\linewidth}
		\caption{Treatment of the function value \ac{SC} for the \gaussian benchmark problem.}
		\label{table:gaussiandomaincons}
		\centering
		\begin{tabular}{@{}p{0.4\textwidth}@{}p{0.35\textwidth}@{}p{0.15\textwidth}}
			\toprule
			\textbf{$\theta$ sample parameters}       & \textbf{$\sigma$ sample parameters}          & \textbf{\ac{SC}}
			\\
			\midrule
			logarithmically in $[-100, -0.01]$        & uniformly in $[0.5, 6]$                      & $f \ge 0$
			\\
			logarithmically in $[0.01, 100]$          & uniformly in $[0.5, 6]$                      & $f \ge 0$
			\\
			\bottomrule
		\end{tabular}
	\end{subtable}
	\begin{subtable}{\linewidth}
		\caption{Treatment of the monotonicity \ac{SC} for the \gaussian benchmark problem.}
		\label{table:gaussianmonocons}
		\centering
		\begin{tabular}{@{}p{0.4\textwidth}@{}p{0.35\textwidth}@{}p{0.15\textwidth}}
			\toprule
			\textbf{$\theta$ sample parameters} & \textbf{$\sigma$ sample parameters} & \textbf{\ac{SC}}
			\\
			\midrule
			logarithmically in $[-0.01, -100]$  & uniformly in $[0.5, 6]$             & $\partial f / \partial \theta \ge 0$
			\\
			logarithmically in $[0.01, 100]$    & uniformly in $[0.5, 6]$             & $\partial f / \partial \theta \le 0$
			\\
			$\{0\}$                             & uniformly in $[0.5, 6]$             & $\partial f / \partial \sigma \le 0$
			\\
			\bottomrule
		\end{tabular}
	\end{subtable}
	\begin{subtable}{\linewidth}
		\centering
		\caption{Treatment of the function value \ac{SC} for the \magman benchmark problem.}
		\label{table:magmandomaincons}
		\begin{tabular}{@{}p{0.4\textwidth}@{}p{0.35\textwidth}@{}p{0.15\textwidth}}
			\toprule
			\textbf{$x$ sample parameters} & \textbf{$I$ sample parameters} & \textbf{\ac{SC}}
			\\
			\midrule
			logarithmically in $[-1000, -0.1]$  & uniformly in $[0.1, 0.8]$           & $f \ge 0$
			\\
			logarithmically in $[0.1, 1000]$    & uniformly in $[0.1, 0.8]$           & $f \le 0$
			\\
			\bottomrule
		\end{tabular}
	\end{subtable}
	\begin{subtable}{\linewidth}
		\centering
		\caption{Treatment of the monotonicity \ac{SC} for the \magman benchmark problem, where $r_1$ and $r_2$ are the roots of the first derivative of the \magman expression.}
		\label{table:magmanmonocons}
		\begin{tabular}{@{}p{0.4\textwidth}@{}p{0.35\textwidth}@{}p{0.15\textwidth}}
			\toprule
			\textbf{$x$ sample parameters}   & \textbf{$I$ sample parameters} & \textbf{\ac{SC}}
			\\
			\midrule
			logarithmically in $[-10, r_1]$  & uniformly in $[0.1, 0.8]$      & $ \partial f / \partial x \ge 0$
			\\
			uniformly in $[r_1, r_2]$        & uniformly in $[0.1, 0.8]$      & $ \partial f / \partial x \le 0$
			\\
			logarithmically in $[r_2, 10]$   & uniformly in $[0.1, 0.8]$      & $ \partial f / \partial x \ge 0$
			\\
			logarithmically in $[-50, -0.1]$ & uniformly in $[0.1, 0.8]$      & $ \partial f / \partial I \ge 0$
			\\
			logarithmically in $[0.1, 50]$   & uniformly in $[0.1, 0.8]$      & $ \partial f / \partial I \le 0$
			\\
			\bottomrule
		\end{tabular}
	\end{subtable}
\end{table}

\subsubsection{Magnetic Manipulator Force}

The second expression we use in our experiments is also used by \cite{KubalikDernerBabuska:2020:1} for their study of \ac{SR} with \ac{SC}.
It describes the force exerted by an electromagnet on an iron ball moving along a rail in a magnetic manipulator system (\magman).
The expression is
\begin{equation}
	\label{eq:magman}
	F
	=
	\alpha \cdot \frac{x \cdot I}{(x^2 + \beta)^3}
	,
\end{equation}
where the independent variables $x$ and $I$ denote the distance and the current, and the parameters $\alpha$ and $\beta$ are system specific parameters.
For $(\alpha, \beta)$, we adopt values $(5.25, 1.75)$ identified by \cite{DamsteegNageshraoBabuska:2017:1}.
The resulting function \eqref{eq:magman} is shown in \Cref{fig:magmanfunc}.
The ground truth expression has a complexity of 11.

For the distance~$x$, the variable range for the fitting data set is $[-3, 3]$, while we take $[-6, 6]$ for the verification data set.
The ranges for the current~$I$ are $[0.1, 0.8]$ and $[0.1, 1.6]$ for the fitting and the verification data set, respectively.
The function set is composed of \parameter{[+,  -,  *,  /,  ^, pow2, pow3]}, where \parameter{pow3} is a unary function raising to the power of~3.

As in the case of the \gaussian problem, function value and monotonicity \ac{SC} are defined for \magman, which are summarized in \Cref{table:magmanmonocons}.
The selection of \ac{SC} evaluation points is the same as described for the \gaussian problem.

\subsubsection{Van der Waals Force}

\begin{figure}[htb]
	\begin{subfigure}[t]{0.48\textwidth}
		\includegraphics[width=1.0\textwidth]{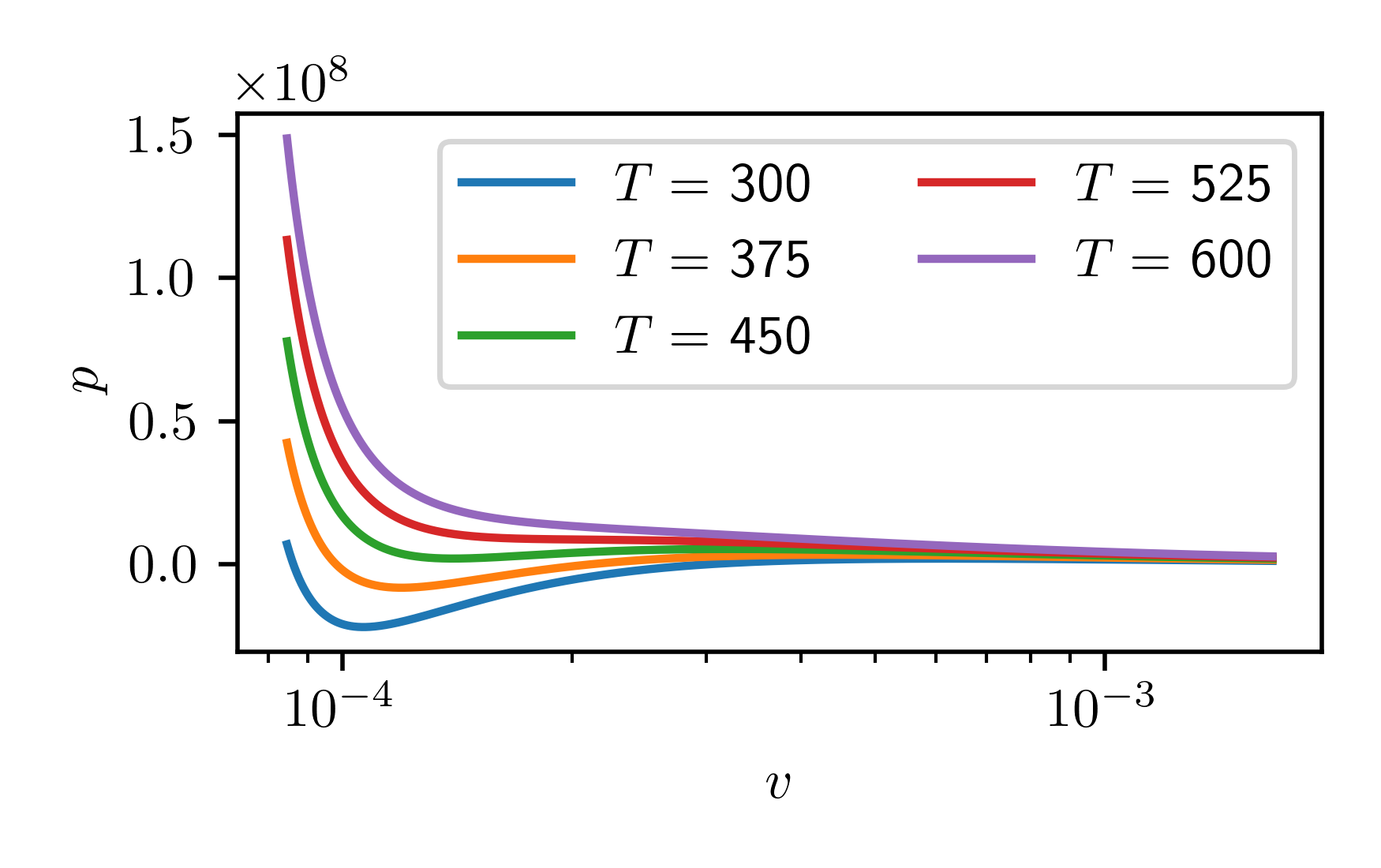}
	\end{subfigure}
	\hfill
	\begin{subfigure}[t]{0.48\textwidth}
		\includegraphics[width=1.0\textwidth]{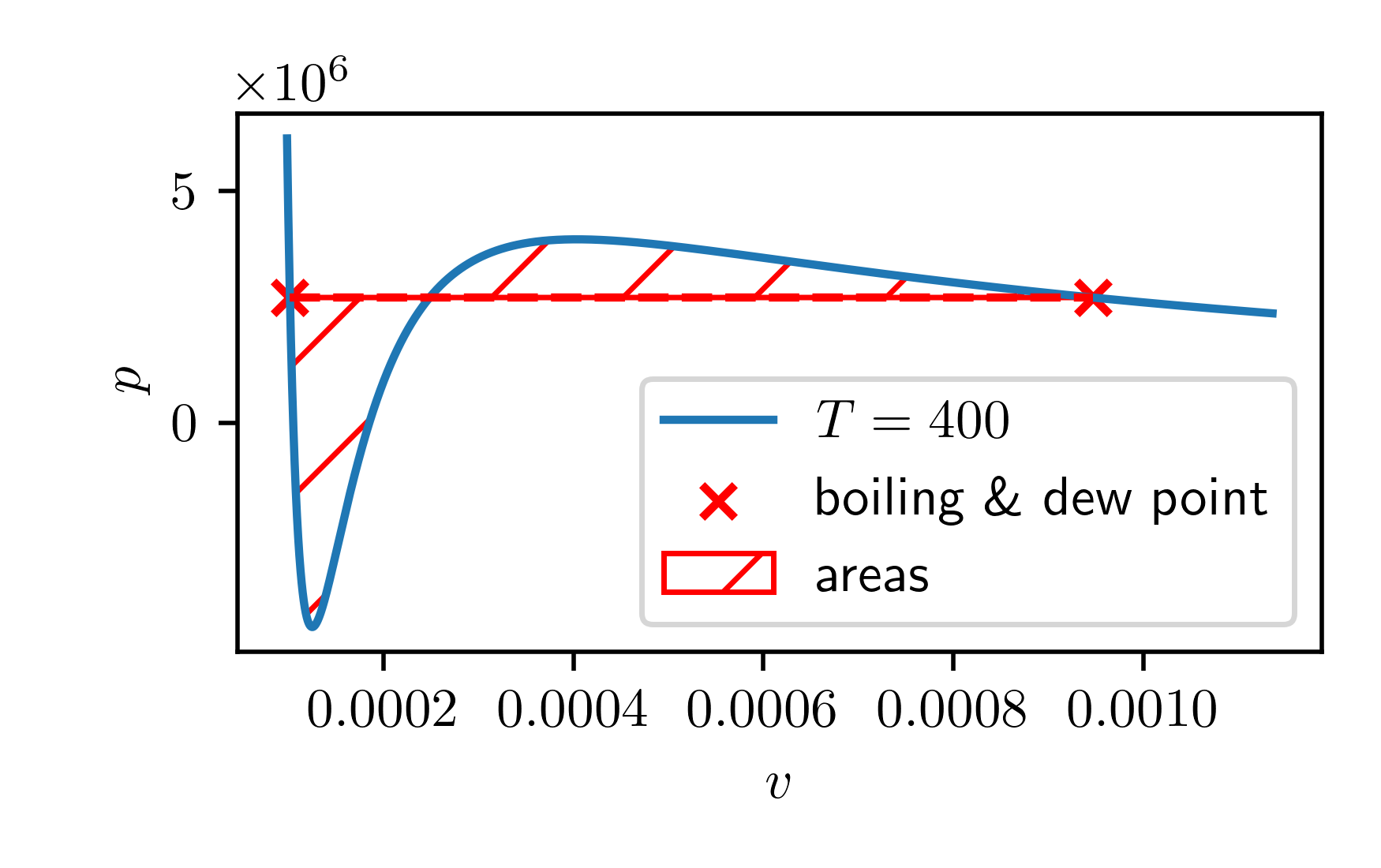}
	\end{subfigure}
	\caption{Plot of the \vanderwaals expression, where the pressure $p$ is shown for varying temperatures $T$ and specific volumes $v$ (left), as well as an illustrative example to aid understanding of the Maxwell criterion at $T = 400$. The boiling and dew points are shown and connected with a dashed red line (right).}
	\label{fig:vanderwaalsfunc}
	\label{fig:maxwellplot}
\end{figure}

The third used expression is the Van der Waals thermodynamic equation of state (\vanderwaals).
It models the pressure~$p$ of a fluid as a function of the molar volume~$v$ and the temperature~$T$ as independent variables:
\begin{equation}
	\label{eq:vanderwaals}
	p
	=
	\frac{R \cdot T}{v - b} - \frac{a}{v^2}
	.
\end{equation}
Here $R \approx \num[round-precision=5]{8.314462618}$ is the universal gas constant, and $a$ and $b$ are fluid-specific parameters.
We use data for methanol, where the parameters $(a, b)$ are $(0.9649, 6.702 \cdot 10^{-5})$ \cite{weast1972handbook}.
\Cref{fig:vanderwaalsfunc} shows the function \eqref{eq:vanderwaals}.
The ground truth expression has a complexity of 12.

We choose the \vanderwaals problem since it involves a relatively complex \ac{SC}, allowing us to demonstrate the capabilities and potential of our approach to incorporate \ac{SC}.
The \ac{SC} for this problem originates from the Maxwell criterion, as it is commonly enforced in thermodynamic equations of state to enable the calculation of the vapor-liquid phase-boundary.
For states $(T,v)$ inside the two-phase region, the pressure calculated by such equations is disregarded.
Instead, the pressure along isotherms is constant throughout the two-phase region.
We briefly summarize the essence to understand the \ac{SC} without going into the thermodynamic detail.
In \Cref{fig:maxwellplot}, the integral of the function along the isotherm (blue line) shifted by the entry points to the two-phase region (dashed red line) must be zero in-between the entry points (red~cross).

We generate three equidistant, regular grids for the \vanderwaals expression, one in the vapor, one in the liquid, and one in the supercritical phase.
To obtain a more realistic grid, the variable ranges are defined to $T$ and $p$ rather than $T$ and $v$, as is common in experimental practice.
The variable ranges are $T = [450, 500]$, $p = [0.05 \cdot 10^6$, $2 \cdot 10^6]$ for the gas phase, $T = [300, 400]$, $p = [6 \cdot 10^6, 7 \cdot 10^6]$ for the liquid phase, and $T = [550, 600]$, $p = [10 \cdot 10^6, 11 \cdot 10^6]$  for the supercritical phase.
Instead of random sampling, the data are sampled equidistantly on regular grids in each of the three ranges.
Additionally, four phase transitions points are added to the fitting data set.
These points are used in the Maxwell criterion \ac{SC}, and thus assumed to be known.
Their values and usage in the \ac{SC} are shown below.
There are 147 and 300~points in the fitting and verification data set, respectively.

The Maxwell criterion is incorporated as an equality constraint.
The exact phase transition states are specified inside the \ac{SC}, marking the numerical integration bounds and the pressure-shifts for two isotherms.
To reduce the computationally expensive numerical integration, the \ac{SC} has two parts.
Before assessing the Maxwell criterion in the second part, the deviations from the phase-transition states are assessed and also treated as equality constraints.
Only for \ac{MARE} to the transition states below $\SI[round-precision=1]{1}{\percent}$, the integration is conducted, the Maxwell criterion checked, and the violation added to the one described above.
To ensure a descent in the transition from the first to the second part of the \ac{SC}, a dummy penalty of 1000 is added to the \ac{MARE}, while the results of the integration are divided by 1000.
This fix does not render the transition smooth, but ensures a descent in this case, and is worth the computational expense saved.
For the isotherms $T_1 = 300$ and $T_2 = 400$, the pressure offsets are $p_1 \approx \num{594598.2419252641}$ and $p_2 \approx \num{2.7042458049626728} \cdot 10^6$.
The specific volumes at the two-phase transitions, which mark the integration bounds, are $v_{1, \mathrm{boiling}} \approx \num{8.609384005897035} \cdot 10^{-5}$ and $v_{1, \mathrm{dew}} \approx \num{0.003847602071128293}$ for $T_1$, and $v_{2, \mathrm{boiling}} \approx \num{0.00010159726806190158}$ and $v_{2, \mathrm{dew}} = \num{0.0009466121805725504}$ for $T_2$.

\subsection{Increasing Noise Levels}

\begin{figure}[htb]
	\includegraphics[width=1.0\textwidth]{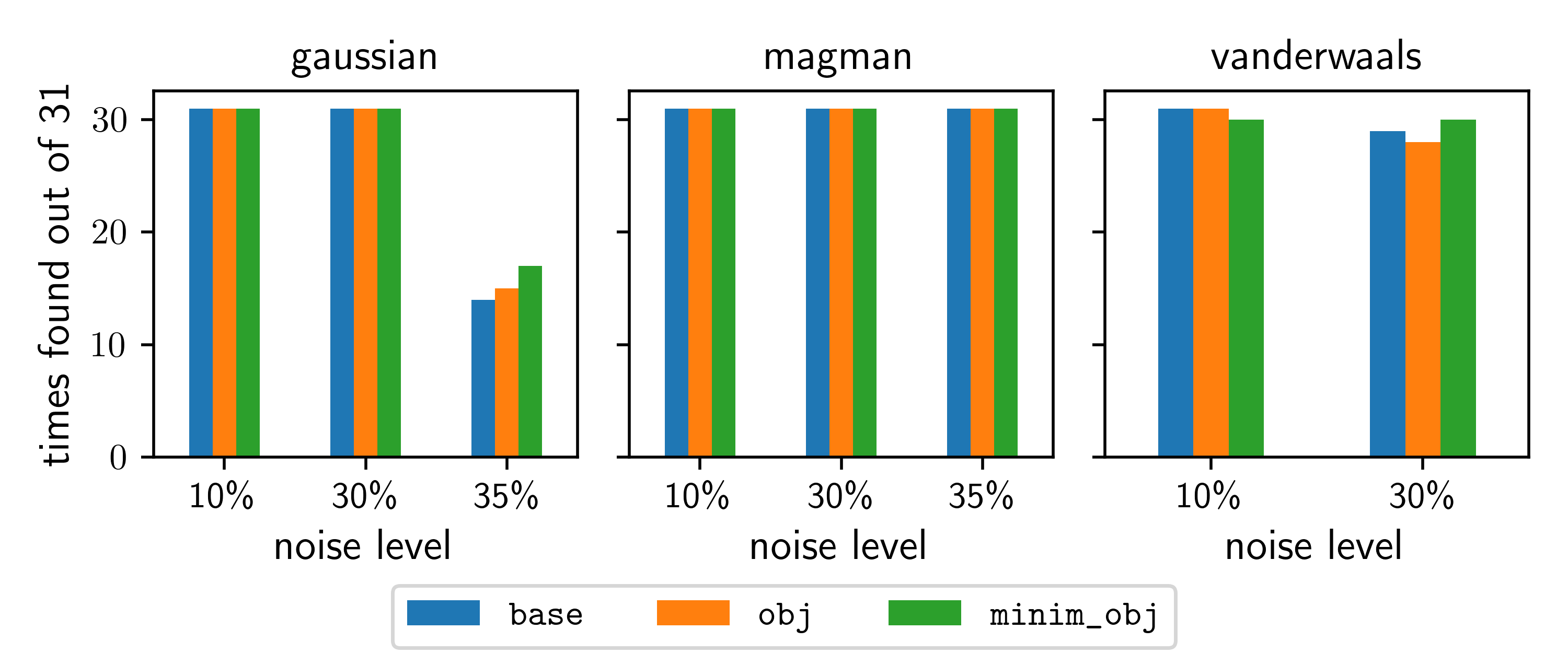}
	\caption{Times out of 31 that each of the three algorithmic variants (\base, \obj, \minimobj) find each of the three ground truth expressions (\gaussian, \magman, \vanderwaals) for noise levels of \SI[round-mode=places, round-precision=0]{10}{\percent}, \SI[round-mode=places, round-precision=0]{30}{\percent}, and \SI[round-mode=places, round-precision=0]{35}{\percent} noise levels.}
	\label{fig:increasingnoiseplot}
\end{figure}

We conduct two studies using the three algorithmic variants and the three benchmark problems.
In the first, the resilience to noise is studied by applying varying noise levels on the dependent component of each data set.
We use normally distributed random noise scaled by the respective target to model relative noise.
Each algorithm-noise level combination is run for 31~times for statistical comparison.
The results are shown in \Cref{fig:increasingnoiseplot}.

For \SI[round-mode=places, round-precision=0]{10}{\percent} and \SI[round-mode=places, round-precision=0]{30}{\percent} noise, all variants are successful between 28 and 31~times out of 31~runs.
Also, for the \magman expression search at \SI[round-mode=places, round-precision=0]{35}{\percent} noise level, all variants find the expression in all runs.
At \SI[round-mode=places, round-precision=0]{35}{\percent} noise for the \gaussian expression, \base finds the expression only 14~times, while \obj and \minimobj find it 15 and 17~times, respectively.
The two-proportion z-test with a confidence level of \SI[round-mode=places, round-precision=0]{5}{\percent} is used to determine whether there are statistically significant differences for the three algorithms.
For this study, the three algorithmic variants do not exhibit statistically significant differences.

\subsection{Reducing Data}

In the first part of the second study, we successively reduce the data available for fitting at $\SI[round-mode=places, round-precision=0]{10}{\percent}$ noise level.
We filter the data sets, by removing the data closest to the normalized center of the data, thus creating a \enquote{data hole}.
First, data is normalized by dividing each datum $x_{ij}$ by the maximum datum of its column $j$:
\begin{equation}
	x_{ij}^{\text{norm}}
	=
	\frac{x_{ij}}{\max(x_{j})}
\end{equation}

Thereafter, the normalized center $\mu_j$ for each column $j$ is determined using
\begin{equation}
	\mu_j
	=
	\frac{1}{n} \sum_{i=1}^{n} x_{ij}^{\text{norm}}
	.
\end{equation}

\begin{figure}[htb]
	\begin{subfigure}[t]{1.0\textwidth}
		\includegraphics[width=1.0\textwidth]{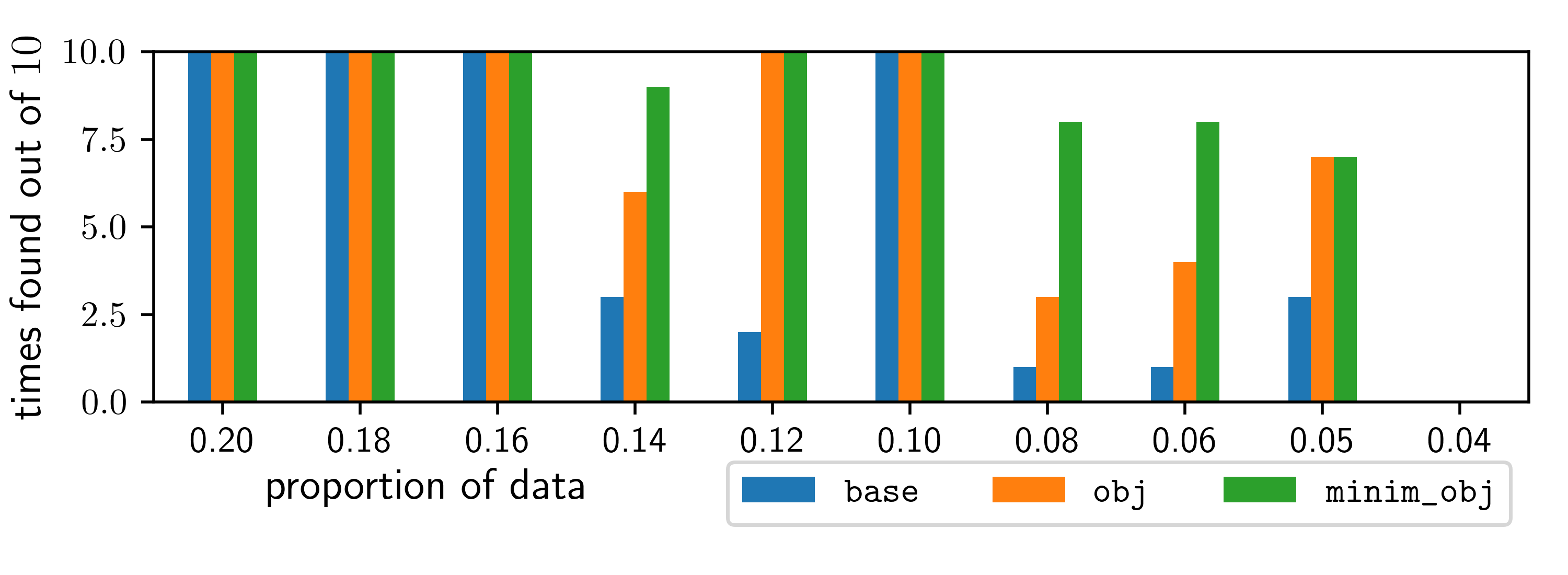}
	\end{subfigure}
	\hfill
	\begin{subfigure}[t]{1.0\textwidth}
		\includegraphics[width=1.0\textwidth]{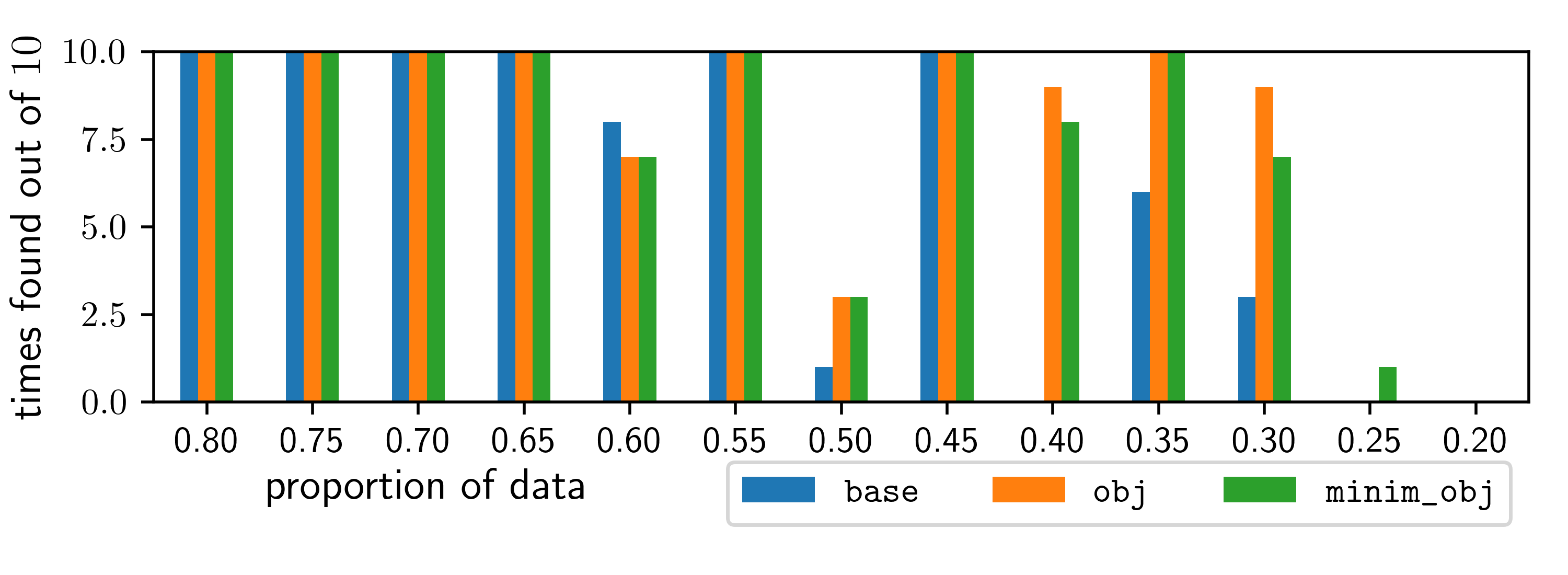}
	\end{subfigure}
	\caption{Times out of ten that each algorithmic variant finds the \gaussian (upper plot) and the \magman (lower plot) expressions at \SI[round-mode=places, round-precision=0]{10}{\percent} noise level for different proportions of data out of 100~data points.}
	\label{fig:reducingdatagaussian}
	\label{fig:reducingdatamagman}
\end{figure}

The data with the highest Euclidean distance to the normalized center, calculated by
\begin{equation}
	d_i^2
	=
	\sum_{j=1}^{m} (x_{ij}^{\text{norm}} - \mu_j)^2
	,
\end{equation}
is removed last.
Each experiment is repeated ten times.
The study is conducted for the \gaussian and the \magman expression.
The results are shown in \Cref{fig:reducingdatagaussian}.

First, the significant results for the \gaussian benchmark problem (see upper plot of \Cref{fig:reducingdatagaussian}) are discussed.
Assuming a significance level of \SI[round-precision=1]{5}{\percent}, for 14~points, \minimobj is statistically significantly better than \base, but not statistically significantly better than \obj.
Also, \obj is not statistically significantly better than \base in this case.
At twelve points, both \obj and \minimobj are statistically significantly better than \base.
For eight points, \minimobj is statistically significantly better than the other two variants, while \obj is not significantly better than \base.
For six points, the only statistically significant result is that \minimobj is better than \base.
At five and four points, none of the results are statistically significant.

The bottom plot of \Cref{fig:reducingdatamagman} shows the same experiments for the \magman expression.
For 40 and 35~points, \obj and \minimobj are significantly better than \base, while not being statistically significantly different.
At 30~points, \obj is significantly better than \base, while \minimobj is not.
However, \obj is not statistically significantly better than \minimobj.

\begin{figure}[hbt]
	\includegraphics[width=1.0\textwidth]{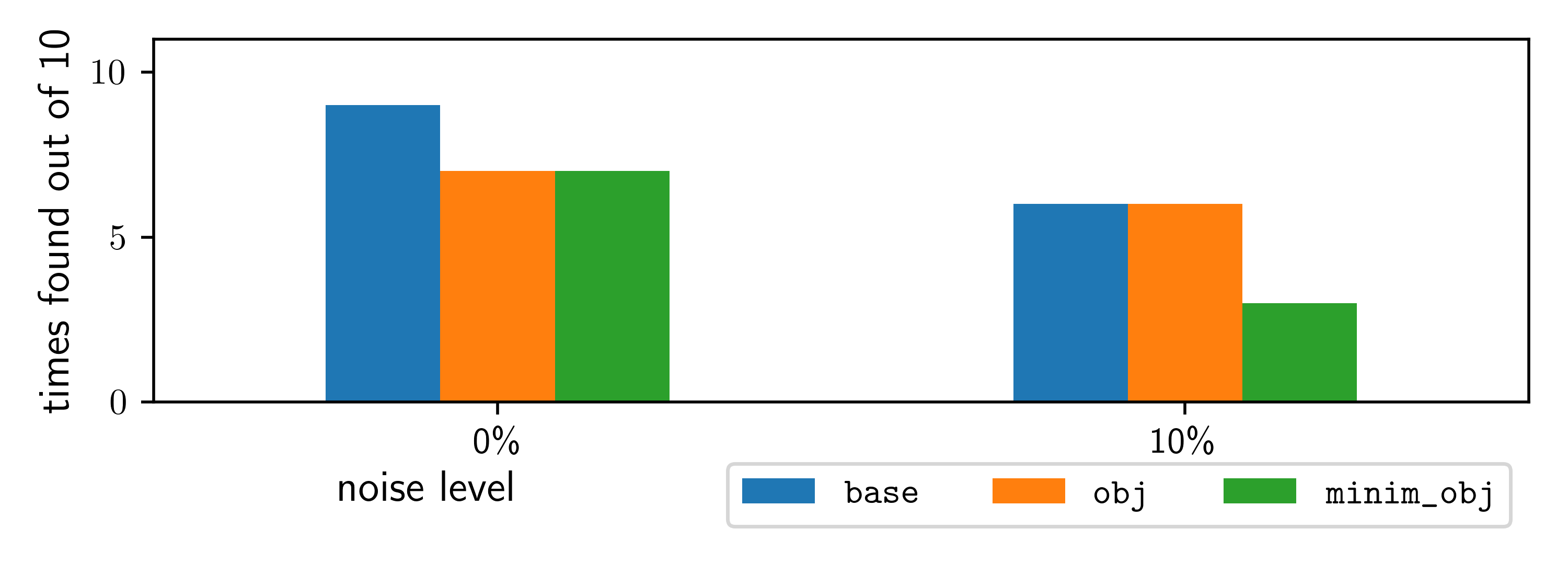}
	\caption{Times out of ten that each algorithmic variant finds the \vanderwaals expression using only data from the liquid phase and four phase transition points at \SI[round-mode=places, round-precision=0]{0}{\percent} and \SI[round-mode=places, round-precision=0]{10}{\percent} noise levels.}
	\label{fig:reducingdatavanderwaals}
\end{figure}

For the \vanderwaals expression, the data are reduced in a different manner.
As mentioned above, the initial \vanderwaals data set consists of three regular grids in three phases, \ie, the gas, the liquid, and the supercritical phase, as well as four points at the vapor and dew line, where no noise is added.
Here, only the data in the liquid phase and the additional points are kept.
Ten experiments are conducted with each, \SI[round-mode=places, round-precision=0]{0}{\percent} and \SI[round-mode=places, round-precision=0]{10}{\percent} noise levels.
The results are shown in \Cref{fig:reducingdatavanderwaals}.
None of the results show statistically significant differences.

\section{Summary and Discussion}
\label{section:discussion}

In \cref{section:experiments}, we compare three algorithmic variants that differ in the way they incorporate used-defined \ac{SC}
\begin{enumeratearabic*}
	\item
		The first baseline variant (\base) does not consider \ac{SC} at all.

	\item
		The second baseline variant (\obj) likewise uses unconstrained least squares for the parameter identification subproblem and subsequently uses the \ac{SC} violation for the purpose of selection.
		This resembles the algorithm proposed by Haider~et~al.\ (see \cite{HaiderOlivettiDeFrancaBurlacuKronberger:2021:1}).

	\item
		The third, proposed variant (\minimobj) already incorporates the \ac{SC} violation into the parameter optimization step.
\end{enumeratearabic*}
Our implementation of all three variants are based on \ac{SR} library \ac{TiSR}.

Several experiments are conducted to test the resilience to noise and to a reduction of the amount of data.
The data is synthetically generated from ground truth expressions.
The experimental setup, including the verification step, allows us to evaluate the differences between the three algorithmic variants with respect to their capability of recovering the ground truth models exactly.

With regard to the noise level, our experiments indicate that the algorithmic variants that utilize \ac{SC} (\obj and \minimobj), do not show significant benefits compared to \base.
With regard to the amount of data, however, the two variants \obj and \minimobj do exhibit statistically significant benefits compared to \base.
Among the two, our proposed approach \minimobj outperforms \obj in a few cases with statistical significance.

Admittedly, we expected our proposed approach to stand out more clearly.
This might be due to several reasons.
For one, for all variants (including \base), \ac{TiSR} employs regularization by adding a squared $\ell_1$-norm of the parameter vector, which makes it more robust against increasing noise levels and overfitting.

Furthermore, the data is synthetically generated using ground truth expressions with normally distributed noise and no outliers.
We conjecture that for problems with systematic deviations or outliers in the data, minimizing \ac{SC} violations during parameter identification might exhibit more significant benefits.

Finally, in situations where data is limited to a relatively small region, but additional knowledge is available concerning some global properties, \ac{SC} are expected to be beneficial to achieve the proper extrapolation behavior.
This situation, however, is not reflected in our current choice of example problems.

\section{Conclusion}
\label{section:conclusion}

In this work, we propose a new way of incorporating \ac{SC} into \ac{SR}.
Our approach uniquely accounts for \ac{SC} violations during the parameter optimization phase and leverages second-order optimization along with algorithmic differentiation to enhance performance.
For this purpose, we evaluate the \ac{SC} violations in a number of sample points and accumulate them into a scalar value.
This value is used both as a penalty during parameter optimization as well as a separate objective during the selection process.
To manage the computational expense, we propose to include the penalty term in to the parameter optimization step only for models that demonstrate good fit quality already in the absence of \ac{SC}.
This preferred variant of our algorithm is referred to as \minimobj.
In our experiments, this approach is statistically significantly better in a few of our test cases compared to two baseline variants, while it is never significantly worse.

Potential future work may study the benefits of this approach for empirical data in the absence of a ground truth model.